\UseRawInputEncoding
\documentclass[conference]{IEEEtran}
\IEEEoverridecommandlockouts
\usepackage{cite}
\usepackage{amsmath,amssymb,amsfonts}
\usepackage{algorithmic}
\usepackage{graphicx}
\usepackage{textcomp}
\usepackage{xcolor}
\IEEEoverridecommandlockouts
\usepackage{eso-pic}

\usepackage{booktabs}
\usepackage{url}
\usepackage{subcaption} 

\def\BibTeX{{\rm B\kern-.05em{\sc i\kern-.025em b}\kern-.08em
    T\kern-.1667em\lower.7ex\hbox{E}\kern-.125emX}}
\begin{document}

\title{MultiHuSE: A Multimodal Dataset for Humour Styles and Emotions
\thanks{This research was supported by the Petroleum Technology Development Fund (PTDF) of Nigeria.}
}

\author{\IEEEauthorblockN{1\textsuperscript{st} Mary Ogbuka Kenneth}
\IEEEauthorblockA{\textit{Algorithmic Human Development group,} \\
\textit{Department of Computing} \\
\textit{Imperial College London}\\
London, United Kingdom \\
m.kenneth22@imperial.ac.uk}
\and
\IEEEauthorblockN{2\textsuperscript{nd} Foaad Khosmood}
\IEEEauthorblockA{\textit{Computer Engineering Department} \\
\textit{California Polytechnic State University}\\
San Luis Obispo, United States \\
foaad@calpoly.edu}
\and
\IEEEauthorblockN{3\textsuperscript{rd} Abbas Edalat}
\IEEEauthorblockA{\textit{Algorithmic Human Development group,} \\
\textit{Department of Computing} \\
\textit{Imperial College London}\\
London, United Kingdom \\
a.edalat@imperial.ac.uk}
}

\AddToShipoutPictureBG*{%
\AtPageLowerLeft{%
\raisebox{1.2cm}{\hspace{1.5cm}%
\parbox{1.0\textwidth}{\footnotesize
\copyright~2025 IEEE. Personal use of this material is permitted. Permission from IEEE must be obtained for all other uses, in any current or future media, including reprinting/republishing this material for advertising or promotional purposes, creating new collective works, for resale or redistribution to servers or lists, or reuse of any copyrighted component of this work in other works. \\[3pt]
This is the accepted version of the paper published as: M. Ogbuka Kenneth, F. Khosmood and A. Edalat, ``MultiHuSE: A Multimodal Dataset for Humour Styles and Emotions,'' 2025 International Conference on Content-Based Multimedia Indexing (CBMI), Dublin, Ireland, 2025, pp. 1--7, doi: 10.1109/CBMI66578.2025.11339313.
}}}}

\maketitle

\begin{abstract}
    Computational recognition of verbal humour remains a challenging task, requiring an understanding of language, delivery style, emotions, and cultural context. Most existing approaches focus on binary classification and lack datasets that capture psychological dimensions of humour alongside variations in expression.
    We introduce MultiHuSE, a multimodal dataset comprising 2,407 high-definition videos of 50 demographically diverse actors performing 1,463 text samples across four psychological humour styles (affiliative, aggressive, self-enhancing, and self-deprecating), as well as neutral content. A subset is additionally annotated for underlying emotions. The dataset uniquely captures multiple actor interpretations of the same texts, enabling systematic analysis of expressive diversity.
   Baseline experiments show that multimodal fusion outperforms unimodal approaches (80.1\% vs. 77.4\% accuracy) in humour style classification, with particularly strong gains for affiliative humour (66\% to 74\%). While text provides the strongest individual signal, fusion models deliver meaningful improvements. We hope that MultiHuSE provides empirical support for psychological theories linking humour and emotion, while also opening new avenues for research in human communication, well-being, and AI-driven interaction. The dataset is available for academic use under an End-User Licence Agreement.
\end{abstract}


\section{Introduction}
Humour recognition presents significant challenges for artificial intelligence systems, particularly in multimodal contexts that require integration of verbal, visual, and acoustic cues \cite{Strapparava2011ComputationalHumour, Kenneth2024ARecognition, Shani2021HowAchievements}. While research in multimodal humour detection has advanced \cite{Kayatani2021TheVideo,Patro2021MultimodalSitcoms,Castro2019TowardsPaper}, most approaches remain limited to binary classification (humour vs. non-humour) \cite{Kenneth2024SystematicClassification}, overlooking deeper psychological dimensions of humour identified by Martin et al. \cite{Martin2003IndividualQuestionnaire}: self-enhancing, self-deprecating, affiliative, and aggressive styles. These categories have significant implications for psychological well-being: affiliative and self-enhancing styles generally foster positive outcomes \cite{Edalat2023Self-initiatedLaugh,Hampes2007TheEmpathy,Plessen2020HumorTraits,Kenneth2025ExplainingAnalysis}, while aggressive and self-deprecating styles can be detrimental to mental health\cite{Veselka2010RelationsPersonality,Khramtsova2016MindfulnessRussia,Kuiper2009HumorWell-being}.

The growing adoption of laughter-based interventions in therapeutic settings for addressing depression and anxiety \cite{Yim2016TherapeuticReview, Akimbekov2021LaughterAnxiety}, combined with evidence that specific humour styles differentially impact mental health outcomes \cite{Ford2017ManipulatingAnxiety, Menendez-Aller2020HumorDepression, Martin2003IndividualQuestionnaire}, underscores the need for computational systems that detect humour styles and their associated emotional expressions. Unlike conventional emotion datasets capturing well-defined emotional states, our approach examines nuanced emotional undertones within humour delivery, emotions that may be subtly expressed or filtered through the comedic performance context \cite{Ritchie2018TheFramework}. This integration enables computational validation of psychological theories and the development of emotionally-aware systems for mental health applications.

Existing multimodal humour datasets present several limitations for analysing humour styles. UR-FUNNY \cite{KamrulHasan2019UR-FUNNY:Humor} focuses on binary humour detection using TED talks; MHD \cite{Patro2021MultimodalSitcoms} uses sitcom recordings with laughter tracks; and Passau-SFCH \cite{Christ2022MultimodalResults} provides style annotations but is restricted to German football press conferences. Additionally, these datasets typically capture only a single performance per instance, limiting the study of expressive variation.

To address these gaps, we introduce \textbf{MultiHuSE}, a new multimodal dataset of over 2,400 recordings featuring fifty diverse performers executing humour across four psychological categories and neutral content, with annotations of underlying emotion. The dataset uniquely captures multiple interpretations of the same texts, enabling systematic analysis of expression variation and multimodal humour perception. Our contributions include:

\begin{enumerate}
    \item The first English-language multimodal dataset annotated with psychological humour styles and enriched with emotion labels.
    \item Multiple video performances of the same text by different actors, enabling analysis of expressive and subjective variation in humour delivery
    \item Benchmark experiments showing that multimodal fusion outperforms unimodal models (80\% vs. 77\% F1-score)
\end{enumerate}

\section{Related Work}
Recent datasets have advanced multimodal humour detection, but with significant limitations (see Table~\ref{tab:exisiting_datasets} for comparison). UR-FUNNY \cite{KamrulHasan2019UR-FUNNY:Humor} contains 16,514 video segments from TED Talks with binary classification, while MHD \cite{Patro2021MultimodalSitcoms} provides 13,633 TV sitcom clips with laughter-based annotations. Both datasets offer valuable scale and naturalistic settings but focus solely on binary humour detection, overlooking psychologically grounded humour styles that have different implications for well-being \cite{Martin2003IndividualQuestionnaire, Kenneth2024ARecognition}.

The Passau-SFCH dataset \cite{Christ2022MultimodalResults} addresses this gap by incorporating four humour styles across 39,682 German football press conference segments, though only 6\% contains actual humour content. However, it is limited to male German speakers in a single domain and lacks controlled recording conditions, limiting its generalisability across diverse populations and contexts. Similarly, YouTube stand-up datasets \cite{Kuznetsova2024MultimodalVideos} capture authentic performances but focus on laughter detection rather than humour style classification.

Multimodal emotion datasets provide methodological foundations for affective analysis but are limited for humour research. MELD \cite{Poria2019MELD:Conversations} includes 13,708 video segments from \textit{Friends} with rich emotion annotations, but only 38\% contain humorous content and lacks humour-specific style labels.   SHEMuD \cite{SinghChauhan2022ASetting} combines binary humour detection with emotion labels across 6,191 TV dialogue segments, but its humour classification remains binary without distinguishing psychologically meaningful humour styles. Both datasets capture only single expressions of content, preventing analysis of how performers might express the same humorous material with varying emotional undertones.

\begin{table*}
  \caption{Comparison of existing multimodal humour and emotion datasets. The table shows key characteristics, including data source, number of video samples, average video duration in seconds, modalities used, annotation granularity, and whether datasets capture single or multiple performances of the same content}
  \label{tab:exisiting_datasets}
  \resizebox{1.0\textwidth}{!}{
  \renewcommand{\arraystretch}{1.3}
  \begin{tabular}{l|l|l|l|l|l|l|l}
    \hline
    \textbf{Dataset} & \textbf{Content Source} & \textbf{\# Videos} & \textbf{Avg Dur.[s]} &\textbf{ Modalities} & \textbf{Annotation Type} & \textbf{\begin{tabular}[c]{@{}l@{}} Performance \\ Type\end{tabular}} &\textbf{ Key Limitations/Features} \\
    \hline
    UR-FUNNY \cite{KamrulHasan2019UR-FUNNY:Humor} & TED Talks & 16,514 & 4.58 & Video, audio, text & Binary humour & Single & English; No humour style differentiation \\
    \hline
    MHD~\cite{Patro2021MultimodalSitcoms} & \begin{tabular}[c]{@{}l@{}} TV sitcoms\\ (Big Bang Theory)\end{tabular} & 13,633 & 3.83 & Video, audio & Binary humour & Single & English; Producer-added laughter; no styles  \\ 
    \hline
    Passau-SFCH~\cite{Christ2022MultimodalResults} & Football press conferences & 39,682 & N/A & Video, audio, text & Four humour styles & Single & German; Limited to male speakers in single domain \\
    \hline
    MELD~\cite{Poria2019MELD:Conversations} & TV series (Friends) & 13,708 & 3.59 & Video, audio, text & Emotions only & Single & English; No humour style annotations  \\
    \hline
    SHEMuD~\cite{Chauhan2022AnDetection} & TV dialogues & 6,191 & N/A & Video, audio, text & Binary humour + emotions & Single & Hindi/English; Single source; no style annotations \\
    \hline
    YouTube Stand-up~\cite{Kuznetsova2024MultimodalVideos} & YouTube stand-up comedy & 39,127 & N/A & Video, audio, text & Laughter-based & Single & Russian/English; Focus on laughter, not humour styles \\
    \hline
    \textbf{MultiHuSE (Ours)} & Controlled studio & 2,407 & 6.66 & Video, audio, text & \textbf{ \begin{tabular}[c]{@{}l@{}} Four humour styles \\ + emotions \end{tabular} }& \textbf{Multiple} & \begin{tabular}[c]{@{}l@{}} English; Diverse performers; consistent quality;\\ \textbf{943 alternative performances of} \\ \textbf{identical jokes with varied expressions} \end{tabular} \\
    
    \hline
  \end{tabular}
  }
\end{table*}

\subsection{Dataset Gaps and Research Motivation}
\label{sec:gaps}
The analysis of existing datasets reveals three critical gaps: (1) predominant focus on binary classification rather than psychologically meaningful humour styles, (2) single-expression capture that overlooks interpretative diversity, and (3) limited integration with psychological theories linking humour to emotional expression and well-being. We developed MultiHuSE with the goal of addressing these gaps and supporting research at the intersection of computation and psychological theory.

\section{Dataset Creation}
We present MultiHuSE, an unpublished multimodal dataset for humour style and emotion recognition. The collection comprises 2,407 high-definition recordings (1080p at 25fps) of fifty demographically diverse participants performing nearly 1,500 text instances. The texts were sourced from the humour style corpus of Kenneth et al. \cite{Kenneth2024ARecognition}, which includes psychologically grounded humour style annotations. While the text annotations are from this prior work, all video recordings, actor performances, and emotion annotations are novel contributions of our dataset.

\subsection{Data Collection}
\paragraph{\textbf{Recording Setup}} We employed a professional, controlled recording environment in a dedicated room at Imperial College London with a Canon EOS Rebel DSLR camera and Rode VideoMic Pro directional microphone. All recordings were conducted under identical setup conditions—white backdrop, consistent lighting, and standardised participant positioning (head to knee)—to ensure both facial expressions and body gestures were clearly visible while minimising environmental variability that could interfere with model training. 

\paragraph{\textbf{Participant Demographics}} We recruited 50 actors (54\% female, 40\% male, 6\% other genders; ages 18--69 years; 72\% White, 12\% Asian/Asian British, 6\% Black/African/Caribbean, 10\% other backgrounds) through a structured recruitment process.\footnote{Detailed recruitment process available at: \url{https://sites.google.com/view/laughterai-com/recruitement}} Participants received a £50 compensation for their $\sim2.5$ hour commitment, including preparation and recording.

\paragraph{\textbf{Participant Preparation}} Participants received detailed instructions explaining each humour style (full instructions available online\footnote{\url{https://sites.google.com/view/laughterai-com/dataset}}), along with the following definitions:
\begin{itemize}
    \item \textit{Affiliative}: Humour used to build connections and social bonds, focusing on shared experiences.
    \item \textit{Self-enhancing}: Humour involving finding amusement in adversity; a positive, adaptive style.
    \item \textit{Aggressive}: Humour used to attack or belittle others through sarcasm or put-downs.
    \item \textit{Self-deprecating}: Humour involving making fun of oneself, highlighting personal weaknesses.
    \item \textit{Neutral}: Statements that are not jokes or do not fit other humour styles.
\end{itemize}

Participants were provided with the text samples prior to the recording day to familiarise themselves with the content, practise delivery with different emotions, and opt out of any samples they felt uncomfortable performing. They were instructed to consider six emotional categories (joy, empathy, anger, neutral, embarrassment, and superiority) while preparing their performances.

\paragraph{\textbf{Dataset Composition}} Our video collection contains over 2,400 samples distributed across five categories with near-balanced representation: self-enhancing (500 videos, 20.8\%), aggressive (529 videos, 22.0\%), neutral (495 videos, 20.6\%), self-deprecating (457 videos, 19.0\%), and affiliative (426 videos, 17.7\%) instances.

\begin{figure}[htbp]
    \centering
    \begin{subfigure}[b]{0.47\columnwidth}
        \includegraphics[width=\textwidth, height=3.2cm]{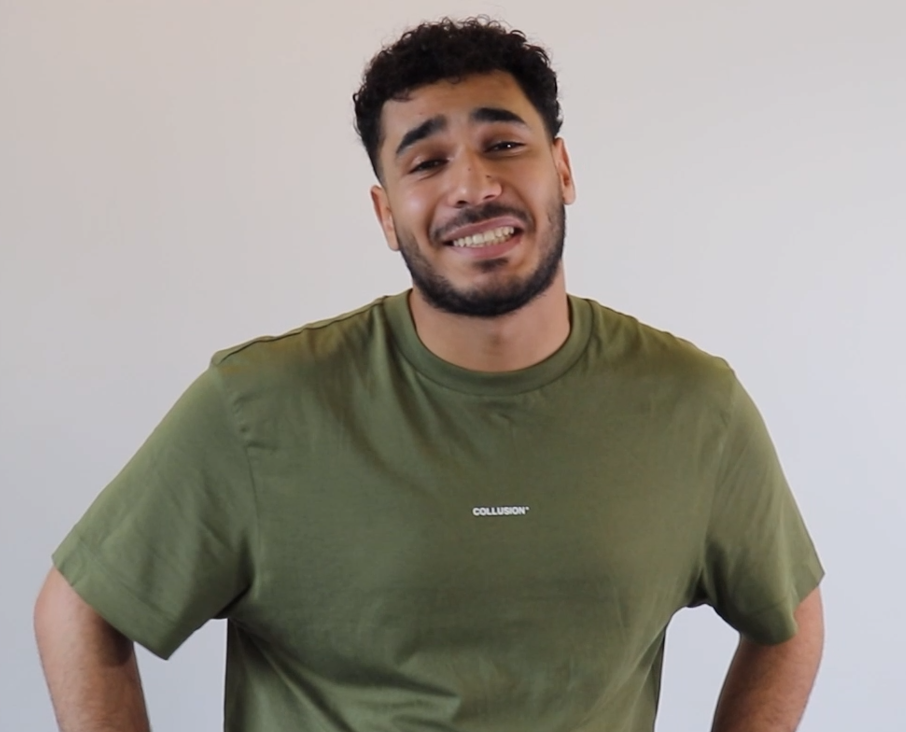}
        \caption{Performer 1 Expressing Anger}
        \label{fig:p2_variability_screenshot_2}
    \end{subfigure}
    \hfill
    \begin{subfigure}[b]{0.47\columnwidth}
        \includegraphics[width=\textwidth, height=3.2cm]{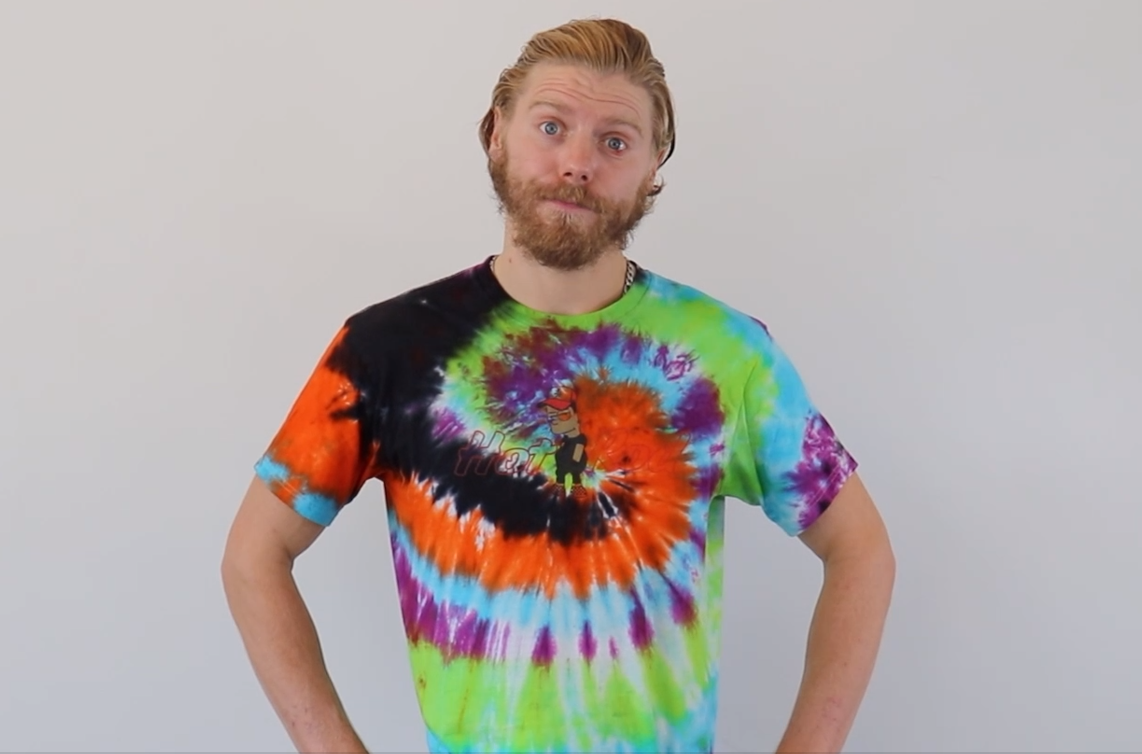}
        \caption{Performer 2 Expressing Superiority}
        \label{fig:p8_variability_screenshot_2}
    \end{subfigure}
    \caption{Expression variability for the same aggressive joke: (a) anger versus (b) superiority delivery of joke: \textit{``I hear you were born on April 2, one day too late"}.}

    \label{fig:performance_variability}
\end{figure}

The dataset uniquely features 943 re-performed videos (approximately 40\% of the dataset), capturing multiple interpretations of identical content for direct comparison of expression variations. This approach enables systematic analysis of the subjective nature of humour delivery, where performers bring unique interpretations to the same textual content. Fig. \ref{fig:performance_variability} illustrates this interpretative diversity through an example of the aggressive humour sample \textit{``I hear you were born on April 2, one day too late''}. The first performer (Fig. \ref{fig:p2_variability_screenshot_2}) delivers the joke with visible anger, whilst the second performer (Fig. \ref{fig:p8_variability_screenshot_2}) expresses a sense of superiority---demonstrating how identical text can elicit distinct emotional expressions and performance styles. Such variation is particularly valuable for studying the nuanced relationship between humour styles, emotional expression, and delivery techniques.

\paragraph{\textbf{Recording Duration}} Videos range from 2 to 74.96 seconds (mean = 6.66\,s), with 91.3\% under 10 seconds, 6.7\% (10--20\,s), 1.2\% (20--30\,s), and 0.8\% ($>30$\,s). Brief videos include concise jokes such as \textit{``Do your thing and don't care if they like it''} (self-enhancing, 2\,seconds), \textit{``They say good things take time so that's why I am always late''} (self-enhancing, 3\,seconds), and \textit{``I feel there is no more trash to burn but myself''} (self-deprecating, 2\,seconds). These brief clips capture complete but concise humorous expressions, while longer videos contain more elaborate joke narratives. This variation in duration reflects the natural diversity of humour delivery, from quick one-liners to extended anecdotes.

\subsection{Annotation and Performance Approach}
\paragraph{\textbf{Humour Style and Text Categories}} We utilised the labels from the existing text dataset~\cite{Kenneth2024ARecognition}, which contains 1,463 instances across five categories: 1,129 jokes aligned with four humour styles framework (self-enhancing, self-deprecating, affiliative, and aggressive)~\cite{Martin2003IndividualQuestionnaire}, and 334 neutral non-jokes.

\begin{figure*}[h]
    \centering
    \includegraphics[width=0.87\textwidth]{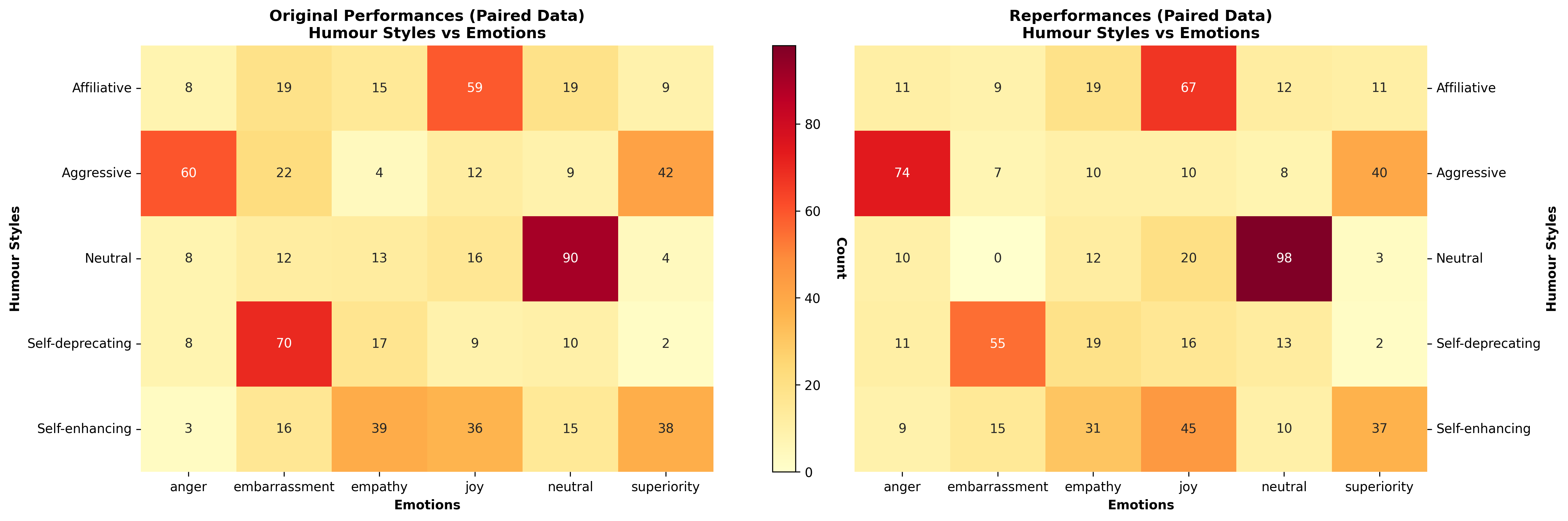}
    \caption{Relationship between humour styles and emotions visualised through heatmaps. Left: original performance annotations; Right: re-performance annotations. Darker colours represent stronger associations between specific emotions and humour styles}
    \label{fig:humour_emotion_heatmap}
\end{figure*}

\paragraph{\textbf{Performance Guidance and Interpretative Freedom}} Actors received a preparation document\footnote{Available at: \url{https://sites.google.com/view/laughterai-com/dataset}} with brief definitions of each humour style, along with the actual jokes and their associated category labels. Crucially, the instructions explicitly stated: \textit{``Remember, there are no `correct' interpretations. We're interested in your natural portrayal of these jokes across different emotional spectrums."} Participants were encouraged to practise delivering each joke with different emotions from a provided list (joy, anger, neutral, superiority, embarrassment, and empathy) to explore varied interpretations. This balanced approach--providing category understanding while emphasising interpretative freedom-- resulted in substantial performance diversity, even for identical text samples, as evidenced by the varying emotions displayed across performances (see Fig.~\ref{fig:humour_emotion_heatmap}, analysed in detail in Section~\ref{subsec:figure3_analysis}).

\paragraph{\textbf{Emotion Annotation}} Immediately after performing each text sample, performers provided self-reports of their own emotional state during delivery, selecting from six emotion categories (joy, anger, neutral, superiority, embarrassment, and empathy). These self-reported emotions represent the performers' internal experience while expressing the humour, not anticipated audience reactions or observer perceptions. This approach aligns with psychological research~\cite{Hampes2007TheEmpathy, Torres-Marin2018IsSpain, Weisfeld2014DoesEmbarrassment, Billig2001HumourLife} that examines humour styles in relation to the expresser's emotional state. These categories enable analysis of humour-embedded emotional states distinct from general emotion datasets, supporting psychological theory validation (Section III-C, Figure 2). This performer-centred emotional assessment provides insight into the psychological experience of humour production rather than reception.

\paragraph{\textbf{Emotion Inter-annotator Agreement}} We evaluated agreement between emotion annotations from original performances and re-performances using paired data where the same jokes were performed by different actors or the same actor with different interpretations. Our analysis used 687 paired instances representing jokes with both original and re-performance emotion annotations.

Our analysis employed two key metrics:
\begin{enumerate}
    \item \textbf{Simple Agreement Rate (0.440)}: The proportion of instances where both annotators assigned the same emotion, calculated as the number of perfect matches (302) divided by the total paired instances (687).
    
    \item \textbf{Cohen's Kappa (0.325)}: A robust measure accounting for chance agreement, indicating fair agreement above random chance~\cite{Sabharwal2021CohensAgreement, Sun2011Meta-analysisKappa}.
\end{enumerate}

While all original performances include emotion annotations, only 687 of the 943 re-performances have emotion labels, allowing for paired analysis of these 687 jokes.

\subsection{Humour Styles-Emotion Relationships}
\label{subsec:figure3_analysis}
Figure~\ref{fig:humour_emotion_heatmap} demonstrates consistent emotion-humour associations across original and re-performances using 687 paired instances. The stable patterns (e.g., aggressive humour predominantly paired with anger/superiority, self-deprecating humour paired with embarrassment) validate that these emotion-humour relationships are robust across different performers, supporting psychological theories that link specific humour styles to corresponding emotional states during expression. Key patterns include:

\begin{itemize}
    \item \textbf{Neutral humour}: Strong association with neutral emotion (Original: 90; Re-performance: 98)
    \item \textbf{Aggressive humour}: Predominantly paired with \textit{Anger} (Original: 60; Re-performance: 74) and \textit{Superiority} (Original: 42; Re-performance: 40), reflecting consistent assertive or mocking interpretations.
    \item \textbf{Self-deprecating humour}: Strongly aligns with \textit{Embarrassment} (Original: 70; Re-performance: 55), supporting its association with vulnerability and self-directed awkwardness.
    \item \textbf{Self-enhancing humour}: Co-occurs with \textit{Empathy} (Original: 39; Re-performance: 31) and \textit{Joy} (Original: 36; Re-performance: 45), indicating reflective and positive self-regard.
    \item \textbf{Affiliative humour}: Primarily linked to \textit{Joy} (Original: 59; Re-performance: 67), supporting its role in positive social bonding.
\end{itemize}

The consistent patterns observed between original and re-performance data indicate emotion-humour associations that align with established psychological theories~\cite{Hampes2007TheEmpathy, Torres-Marin2018IsSpain, Weisfeld2014DoesEmbarrassment, Billig2001HumourLife} and offer potential support for computational approaches to recognising emotional aspects of humour styles.

\subsection{Ethical Considerations and Data Access}
The MultiHuSE dataset was created with careful attention to ethical principles. All procedures received approval from the Science, Engineering and Technology Research Ethics Committee at Imperial College London. Participants provided informed consent, were compensated \pounds50 for their time, and had the option to opt out of performing any content they found uncomfortable. While personal identifiers were pseudonymised, facial features remained visible as required for research purposes, with participants explicitly informed of this necessity.

The MultiHuSE dataset is available to researchers via a dedicated section of our project website\footnote{Dataset available at: \url{https://sites.google.com/view/laughterai-com/dataset}}, under a custom End-User Licence Agreement (EULA) permitting use for academic and research purposes only. The website currently provides information about the project and access procedures.

To mitigate potential risks associated with offensive content in certain aggressive humour samples, the dataset documentation explicitly states that inclusion does not imply endorsement. Use of the dataset is restricted to research purposes aligned with the advancement of mental health understanding and computational humour analysis.

\section{Experiments}
To demonstrate the viability and research utility of our multimodal collection, we performed baseline experiments evaluating humour style and emotion recognition across single and combined modalities. Direct comparison with existing datasets is not feasible as MultiHuSE represents the first English-language multimodal dataset with psychological humour style annotations, making our baselines foundational benchmarks for future research.

\subsection{Feature Extraction}
We employed state-of-the-art encoders to extract robust multimodal features: 

\textbf{Text:} BERT-base-uncased embeddings (768-D), using the [CLS] token from padded sequences (max 128 tokens) \cite{Devlin2019BERT:Understanding}

\textbf{Audio:} Dasheng-0.6B (1280D) \cite{Dinkel2024ScalingClassification}, a powerful audio encoder pretrained on diverse audio data for capturing nuanced speech and sound features.

\textbf{Video:} MC3-18 (768D) \cite{Tran2018ARecognition}, a spatiotemporal convolutional encoder optimised for short video clips and fine-grained motion representation.

\subsection{Data Splits and Preprocessing}
We used an 80:20 train-test split stratified by humour style for humour style classification. Training data underwent 5-fold stratified cross-validation.

\subsection{Baseline Models}
We evaluated XGBoost classifiers and Cross-Modal Attention Transformer:

\textbf{XGBoost:} \texttt{XGBClassifier(use\_label\_encoder =False, random\_state=42, eval\_metric='mlogloss')} \cite{Chen2016XGBoost:System}.

\textbf{Transformer:} Cross-modal attention network with multi-head attention (8 heads) projecting text, audio, and visual features to a common 1024D space, trained end-to-end for humour style and emotions classification.

\subsection{Fusion Strategies}
\textbf{Single Modality:} Individual XGBoost models trained for text, audio, and visual encoder features.

\textbf{Exponential Weighted Fusion:} Predictions from separate XGBoost modality models combined using exponential weighting ($\alpha = 2$) based on individual modality performance, resulting in dynamic weights (humour: text 0.54, audio 0.31, visual 0.15; emotion: text 0.40, audio 0.32, visual 0.27).

\textbf{Cross-Modal Attention:} Uses an attention mechanism to weigh the relevance of features from one modality when processing another. This allows the model to learn complex relationships between modalities directly.

\subsection{Evaluation}
We report accuracy, precision, recall, and F1-score (weighted averages). Accuracy measures overall prediction correctness, while precision reflects the proportion of true positives among predicted positives. Recall assesses the model’s ability to identify all actual positives, and the F1-score provides the harmonic mean of precision and recall, balancing both metrics.

\section{Results}

\subsection{Modality Effectiveness Analysis}
Table~\ref{tab:baseline-eval} shows baseline results. Text features yielded the strongest performance (77.4\%), confirming linguistic content as the primary signal for humour style classification. Audio (58.5\%) and visual (40.0\%) lagged behind, reflecting the subtle and subjective nature of emotional and humorous expression, which makes these cues harder to model. Still, they add complementary prosodic and expressive information.

Multimodal fusion outperformed unimodal baselines: exponential weighted fusion reached 80.1\% accuracy and cross-attention 79.7\%. Though gains over text-only are modest (+2.7\%, +2.3\%), they highlight meaningful cross-modal contributions and demonstrate the dataset’s compatibility with modern architectures.

\begin{table}[htbp]
\caption{Comprehensive Multimodal Results on MultiHuSE Dataset}
\label{tab:baseline-eval}
\centering
\resizebox{0.48\textwidth}{!}{%
\begin{tabular}{lcccccccc}
\toprule
\textbf{Method} & \multicolumn{4}{c}{\textbf{Humor Styles (\%)}} & \multicolumn{4}{c}{\textbf{Emotion (\%)}} \\
\cmidrule(lr){2-5} \cmidrule(lr){6-9}
& Acc & Prec & Recall & F1 & Acc & Prec & Recall & F1 \\
\midrule
\multicolumn{9}{l}{\textit{Single Modality Baselines}} \\
Text Only & 77.4 & 78.0 & 77.0 & 77.0 & 36.5 & 35.0 & 37.0 & 35.0 \\
Audio Only & 58.5 & 59.0 & 58.0 & 58.0 & 32.8 & 33.0 & 33.0 & 32.0 \\
Visual Only & 40.0 & 40.0 & 40.0 & 39.0 & 30.1 & 30.0 & 30.0 & 29.0 \\
\midrule
\multicolumn{9}{l}{\textit{Multimodal Fusion}} \\
Equal Fusion & 77.8 & 78.0 & 78.0 & 78.0 & 38.2 & 38.0 & 38.0 & 37.0 \\
Exponential Fusion & \textbf{80.1} & \textbf{80.0} & \textbf{80.0} & \textbf{80.0 }& \textbf{38.6} & \textbf{38.0} & \textbf{39.0} & \textbf{37.0} \\
Cross-Attention & 79.7 & 80.0 & 79.0 & 79.0 & 32.2 & 32.0 & 32.0 & 32.0 \\
\bottomrule
\end{tabular}
}
\end{table}

Table~\ref{tab:humor_style_results} shows humour-style variation: neutral content was most reliable (85 - 90\% F1), while affiliative humour benefited most from multimodality (66\% to 74\% F1 with exponential fusion). Text remained dominant across all fusions, with audio-visual alone achieving only 58.3\%. Thus, humour style recognition is semantically driven but enriched by multimodal expression, underscoring the capacity of the data set to provide a complete understanding of humour.

\begin{table}[h]
\caption{F1-Scores for Individual Humour Styles}
\label{tab:humor_style_results}
\small
\resizebox{0.48\textwidth}{!}{
\begin{tabular}{lccccc}
\toprule
\textbf{Method} & \textbf{Self-Enhancing} & \textbf{Self-Deprecating} & \textbf{Affiliative} & \textbf{Aggressive} & \textbf{Neutral} \\
\midrule
Text Only & 77 & 80 & 66 & 77 & 85 \\
Audio Only & 55 & 57 & 47 & 54 & 79 \\
Visual Only & 39 & 29 & 33 & 40 & 56 \\
\midrule
Equal Fusion & 76 & 79 & 66 & 76 & 90 \\
Exponential Fusion &\textbf{79}& \textbf{81} &\textbf{74} &\textbf{78} & \textbf{89} \\
Cross-Attention & 83 & 78 & 68 & 81 & 85 \\
\bottomrule
\end{tabular}
}
\end{table}

\subsection{Dataset Quality and Validation}

To assess the reliability and research utility of the dataset, we conducted validation across three key dimensions: feature extraction reliability, re-performance variation, and annotation coverage.

\subsubsection{Feature Extraction Reliability}
To validate technical data quality, we assessed computational feature extraction success rates across all video samples using state-of-the-art encoders. Table~\ref{tab:extraction_rates} presents the success rates for each modality.

\begin{table}[h]
\centering
\caption{Feature extraction success rates}
\label{tab:extraction_rates}
\scriptsize
\begin{tabular}{lccc}
\toprule
\textbf{Modality} & \textbf{Encoder} & \textbf{Success Rate} & \textbf{Successful/Total} \\
\midrule
Text    & BERT-base-uncased & 100\%  & 2,407/2,407 \\
Audio   & Dasheng-0.6B   & 100\%   & 2,407/2,407 \\
Visual  & MC3-18 & 100\%  & 2,407/2,407 \\
\bottomrule
\end{tabular}
\end{table}

All modalities achieved perfect extraction rates (100\%) using modern pre-trained encoders, ensuring comprehensive coverage for multimodal fusion experiments.

\subsubsection{Re-performance Variation}
A central strength of our dataset is its capture of diverse interpretations of identical humorous content. We analysed 1,872 re-performance videos corresponding to 927 unique jokes, using 17 facial features extracted via MediaPipe Face Mesh \cite{Kartynnik2019Real-timeGPUs}. For each joke performed by multiple actors, we computed pairwise Euclidean distances between facial feature vectors and compared them to random baseline distances.

\begin{table}[h]
\centering
\caption{Re-performance variation analysis}
\label{tab:reperformance_variation}
\scriptsize
\begin{tabular}{lcc}
\toprule
\textbf{Distance Type} & \textbf{Mean ± SD} & \textbf{Interpretation} \\
\midrule
Within-joke & 0.2181 ± 0.1112 & Same joke re-performances \\
Random baseline & 0.2661 ± 0.1181 & Different jokes \\
\midrule
\textbf{Mann-Whitney U} & \textbf{$p < 0.000001$} & \textbf{Cohen's d = 0.43} \\
\bottomrule
\end{tabular}
\end{table}

Within-joke distances were significantly smaller than random baseline distances, indicating that re-performances of identical content maintain meaningful similarity whilst exhibiting substantial expressive diversity. This statistically significant difference suggests that re-performances capture both content consistency and interpretative variation, supporting MultiHuSE's ability to study the subjective nature of humour expression.

\subsubsection{Annotation Coverage}

All 2,407 videos include complete humour style labels inherited from the source text dataset. Self-reported emotion annotations are available for all 1,463 original performances and 687 re-performances (72.7\% of re-performed instances), providing 89.3\% overall emotion coverage (2,150/2,407 instances). The partial emotion coverage resulted from methodological evolution during data collection, as the emotion annotation protocol for re-performance was introduced partway through the study after recognising its value for understanding expression variation.

\section{Applications of the MultiHuSE Dataset}
MultiHuSE enables recognition of psychologically grounded humour styles with established links to mental health outcomes. Meta-analytic evidence shows affiliative and self-enhancing humour protects against depression and anxiety~\cite{Menendez-Aller2020HumorDepression,Ford2017ManipulatingAnxiety}, while aggressive and self-deprecating styles increase mental health risks~\cite{Kuiper2009HumorWell-being,Amjad2022HumorAdults}. Recent laughter-based interventions for treating depression~\cite{Akimbekov2021LaughterAnxiety,Yim2016TherapeuticReview} highlight the therapeutic potential of humour style differentiation.

\subsection{Mental Health Applications}
\textbf{Digital Monitoring}: Apps that track users' humour expressions during therapy sessions--detecting predominant self-deprecating humour (depression risk indicator) versus affiliative humour (protective factor)--enabling personalised intervention recommendations. For example, a system could alert therapists when patients consistently use self-deprecating humour, prompting targeted cognitive behavioural interventions.

\textbf{Clinical Assessment}: Tools supporting mental health professionals in evaluating humour-based coping mechanisms and therapeutic progress through automated analysis of patient expressions.

\textbf{Content Moderation}: Systems detecting harmful humour while preserving adaptive styles, reducing cyberbullying and enhancing digital well-being.

\textbf{Emotionally Aware AI}: Virtual assistants that differentiate supportive from harmful humour, enabling psychologically sensitive responses in therapeutic contexts.

\section{Conclusion}
We present MultiHuSE, a multimodal dataset for humour style and emotion recognition comprising over 2,400 high-definition recordings of fifty diverse performers executing a balanced set of textual content across four psychologically-grounded categories. The dataset includes 943 re-performed samples, enabling direct comparison of expression variations for identical content.

Our validation shows 100\% feature extraction success with state-of-the-art encoders and significant performance variation reflecting interpretative diversity ($p < 0.000001$). Multimodal fusion outperformed single modalities, with exponential fusion achieving 80.1\% accuracy vs.\ 77.4\% for text-only; cross-attention reached 79.7\%, confirming compatibility with modern deep learning. Ablation highlights text dominance while showing complementary contributions from audio and visual cues. We have observed relationships between humour styles and specific emotions that appear consistent with psychological theories.

\section{LIMITATIONS AND FUTURE DIRECTIONS}

We acknowledge several limitations. First, the moderate inter-annotator agreement (Cohen's $\kappa = 0.325$) reflects the inherent subjectivity of emotional experiences in humour contexts. Second, despite demographic diversity within our sample 72\% White, 12\% Asian/Asian British, 6\% Black/African/Caribbean, 10\% other backgrounds), recruitment was geographically constrained to London, which may limit the dataset's representation of global humour expression patterns and cultural interpretations of humour styles. This geographical limitation reinforces Western humour perspectives and may introduce cultural bias in computational models trained on this data.

Third, while our dataset contains 2,407 videos compared to larger collections, our controlled design emphasises quality and balance--79.4\% humorous content with even distribution across styles, compared to Passau-SFCH's 6.02\% humorous content from a single domain. Direct comparison with existing datasets is challenging as MultiHuSE represents the first English-language multimodal dataset with psychological humour style annotations.

Future research will address these limitations through multiple recruitment centers across different countries and cultural contexts to capture diverse humour expression patterns and cultural interpretations of humour styles. Additional directions include exploring advanced vision-language models, longitudinal studies of humour style changes for intervention applications, and investigating relationships between automatically detected and human-perceived emotions in humour contexts.

\bibliographystyle{IEEEtran}
\bibliography{references}

\end{document}